\documentclass[preprint]{sig-alternate-05-2015}

\usepackage{pdf14}
\usepackage[breaklinks=true,bookmarks=false,colorlinks=true]{hyperref}
\usepackage{makecell}
\usepackage{enumitem}
\setlist{nolistsep}

\begin{document}

% Copyright
%\setcopyright{acmcopyright}
%\setcopyright{acmlicensed}
%\setcopyright{rightsretained}
%\setcopyright{usgov}
%\setcopyright{usgovmixed}
%\setcopyright{cagov}
%\setcopyright{cagovmixed}

%\CopyrightYear{2016} 
%\setcopyright{acmcopyright}
%\conferenceinfo{MM '16,}{October 15-19, 2016, Amsterdam, Netherlands}
\isbn{}
%\acmPrice{\$15.00}
\doi{}

% --- End of Author Metadata ---

\title{Enabling My Robot To Play Pictionary : Recurrent Neural Networks For Sketch Recognition}

\numberofauthors{3} %  in this sample file, there are a *total*
% % of EIGHT authors. SIX appear on the 'first-page' (for formatting
% % reasons) and the remaining two appear in the \additionalauthors section.
% %

\author{
      Ravi Kiran Sarvadevabhatla\\      
      \email{ravikiran@grads.cds.iisc.ac.in}
\and
      Jogendra Kundu\titlenote{equal contribution as first author}\\     
      \email{jogendranathkundu@gmail.com}
\and
      Venkatesh Babu R\\    
      \email{venky@cds.iisc.ac.in}
\and
      %\sharedaffiliation
      \affaddr{Video Analytics Lab}\\
       \affaddr{Department of Computational and Data Sciences}\\
       \affaddr{Indian Institute of Science}\\
       \affaddr{Bangalore, INDIA}
}

\maketitle
\begin{abstract}
 Freehand sketching is an inherently sequential process. Yet, most approaches for hand-drawn sketch  recognition either ignore this sequential aspect or exploit it in an ad-hoc manner. In our work, we propose a recurrent neural network architecture for sketch object recognition which exploits the long-term sequential and structural regularities in stroke data in a scalable manner. Specifically, we introduce a Gated Recurrent Unit based framework which leverages deep sketch features and weighted per-timestep loss to achieve state-of-the-art results on a large database of freehand object sketches across a large number of object categories. The inherently online nature of our framework is especially suited for on-the-fly recognition of objects as they are being drawn. Thus, our framework can enable interesting applications such as camera-equipped robots playing the popular party game Pictionary with human players and generating sparsified yet recognizable sketches of objects.
\end{abstract}

\printccsdesc

% We no longer use \terms command
%\terms{Theory}


\section{Introduction}
The process of freehand sketching has long been employed by humans to communicate ideas and intent in a minimalist yet almost universally understandable manner. In spite of the challenges posed in recognizing them~\cite{li2013sketch}, sketches have formed the basis of applications in areas of forensic analysis~\cite{klare2011matching}, electronic classroom systems~\cite{meyer2009intelligent}, sketch-based retrieval~\cite{sun2012query,li2013sketch} etc. 

Sketching is an inherently sequential process. The proliferation of pen and tablet based devices today enables us to capture and analyze the entire process of sketching, thus providing additional information compared to passive parsing of static sketched content. Yet, most sketch recognition approaches either ignore the sequential aspect or lack the ability to exploit it~\cite{sun2012query,eitz2012hdhso,Schneider2014}. The few approaches which attempt to exploit the sequential sketch stroke data do so either in an unnatural manner~\cite{yang2015deep} or impose restrictive constraints (e.g. Markov assumption)~\cite{arandjelovic2011sketch}.  

In our work, we propose a recurrent neural network architecture for sketch object recognition which exploits the long-term sequential and structural regularities in stroke data in a scalable manner. We make the following contributions:
\begin{itemize}
    \item We propose the first deep recurrent neural network architecture which can recognize freehand sketches across a large number ($160$) of  object categories. Specifically, we introduce a Gated Recurrent Unit (GRU)-based framework (Section \ref{sec:overview}) which leverages deep sketch features and weighted per-stroke loss to achieve state-of-the-art results.
    \item We show that the choice of deep sketch features and recurrent network architecture \textit{both}  play a crucial role in obtaining good recognition performance (Section \ref{sec:results}).
    \item Via our experiments on sketches with partial temporal stroke content, we show that our framework recognizes the largest percentage of sketches (Section \ref{sec:results}). 
\end{itemize}

Given the on-line nature of our recognition framework, it is especially suited for on-the-fly interpretation of sketches as they are drawn. Thus, our framework can enable interesting applications such as camera-equipped robots playing the popular party game Pictionary~\cite{pictionary} with human players, generating sparsified yet recognizable sketches of objects~\cite{Sarvadevabhatla2015}, interpreting hand-drawn digital content in  electronic classrooms~\cite{meyer2009intelligent} etc.

\begin{figure*}[!tbp]
    \centering    
		\includegraphics[width=0.7\textwidth,keepaspectratio]{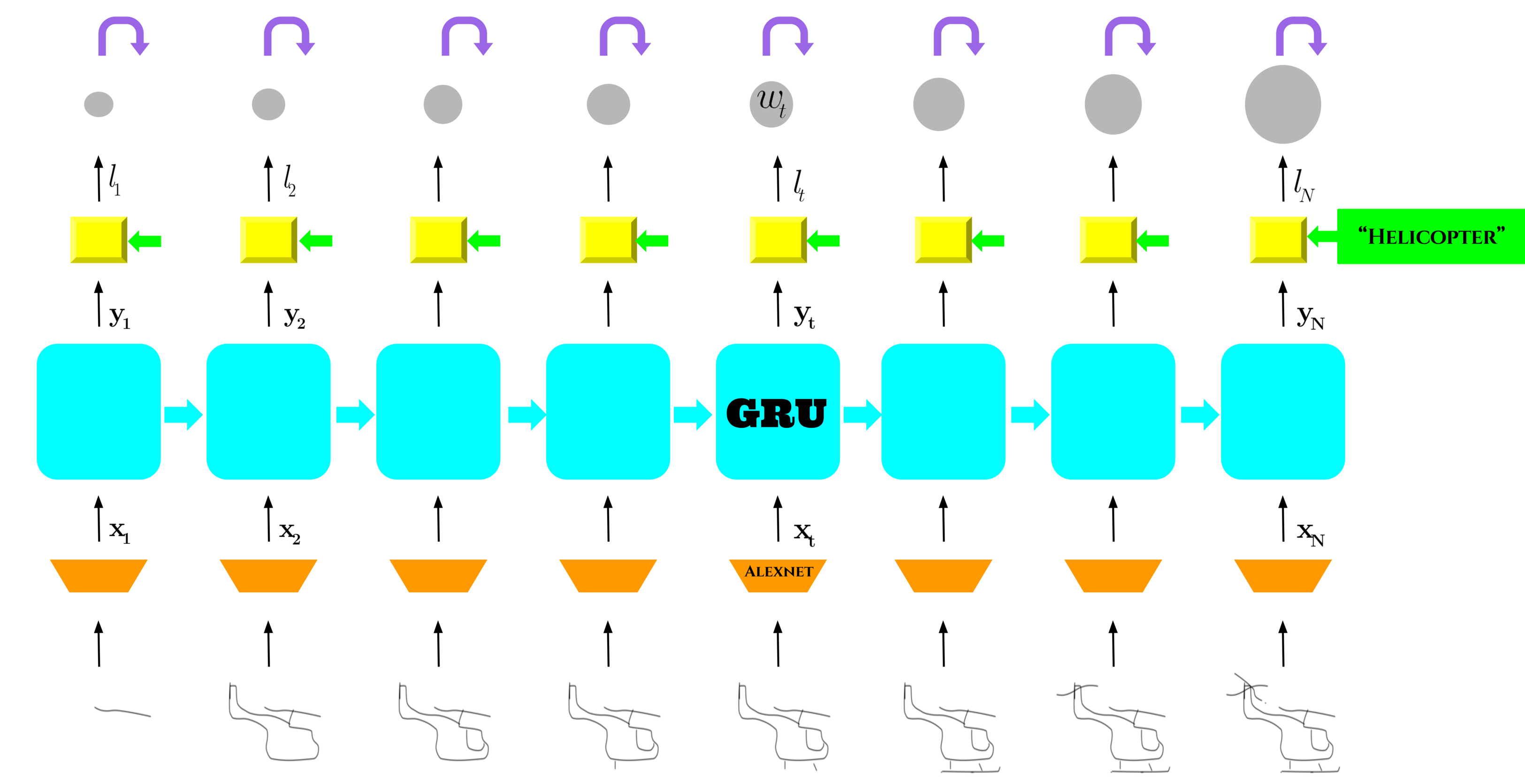}
		\caption{Overview of our sketch-recognition framework. Alexnet-based deep features ($\mathbf{x_{\!_{1}},x_{\!_{2}},}\ldots \mathbf{,x_{\!_{t}},}\ldots $) from each cumulative stroke image form the input to the GRU (blue), obtaining the corresponding prediction sequence ($\mathbf{y_{\!_{1}},y_{\!_{2}},}\ldots \mathbf{,y_{\!_{t}},}\ldots $). The per-timestep loss $l_t$ is computed w.r.t ground-truth (green arrow) by the loss function (yellow box). This loss is weighted by a corresponding $w_{\!_{t}}$ (shown as proportionally sized gray circles) and backpropagated (purple curved arrow) for the corresponding time step $t$. Best viewed in color.} 
		\label{fig:overview}    
\end{figure*}

\vspace{-3mm}

\section{Related Work}
\label{sec:relatedwork}

To retain focus, we review approaches exclusively related to recognition of hand-drawn object sketches.
Early datasets tended to contain either a small number of sketches and/or object categories~\cite{sun2012query,lovett2006}. In $2012$, Eitz et al.~\cite{eitz2012hdhso} released a dataset containing $20000$ hand-drawn sketches across $250$ categories of everyday objects. The dataset, currently the largest sketch object dataset available, provided the first opportunity to attempt the sketch object recognition problem at a relatively large-scale. Since its release, a number of approaches have been proposed to recognize freehand sketches of objects. The initial performance of handcrafted feature-based approaches~\cite{li2013sketch,Schneider2014} has been recently surpassed by deep feature-based approaches~\cite{Sarvadevabhatla2015,seddati2015deepsketch}, culminating in an custom-designed Convolutional Neural Network dubbed SketchCNN which achieved state-of-the-art results~\cite{yang2015deep}.
The approaches mentioned above are primarily designed for static, full-sketch object recognition. In contrast, another set of approaches attempt to exploit the sequential stroke-by-stroke nature of hand-drawn sketch creation~\cite{arandjelovic2011sketch,wang2010line}. For example, Arandjelovic and Sezgin~\cite{arandjelovic2011sketch} propose a Hidden Markov Model (HMM)-based approach for recognizing military and crisis management symbol objects. Although mentioned above in the context of static object recognition, a variant of the SketchCNN~\cite{yang2015deep} can also handle sequential stroke data. In fact, the authors demonstrate that exploiting the sequential nature of sketching process improves the overall recognition rate. However, given that CNNs are not inherently designed to preserve sequential ``state'', better results can be expected from a framework which handles sequential data in a more natural fashion. The approach we present in our paper aims to do precisely this. Our framework is based on Gated Recurrent Unit (GRU) networks recently proposed by Cho et al.~\cite{ChoMGBSB14}. GRU architectures share a number of similarities with the more popular Long Short Term Memory Networks~\cite{hochreiter1997long}  including the latter's ability to perform better~\cite{graves2009novel} than traditional models (e.g. HMM) for problems involving long and complicated sequential structures. To the best of our knowledge, recurrent neural networks have not been utilized for online sketch recognition. 

\vspace{-3mm}

\section{Our recognition framework}
\label{sec:recframework}

\vspace{-2mm}

\subsection{Overview}
\label{sec:overview}

Sketch creation involves accumulation of hand-drawn strokes over time. Thus, we require our recognition framework to optimally exploit object category evidence being accumulated on a per-stroke basis as well as temporally. Moreover, the variety in sketch-based depiction and intrinsic representational complexity of objects results in a large range for stroke-sequence lengths.
%(\textit{Need range here}). 
Therefore, we require our recognition framework to address this variation in sequence lengths appropriately. To meet these requirements, we employ Gated Recurrent Unit (GRU) networks~\cite{ChoMGBSB14}. Our choice of GRU architecture is motivated by the observation that it involves learning a smaller number of parameters and performs better compared to LSTM in certain instances~\cite{chung2014empirical} including, as shall been seen (Section \ref{sec:experiments}), our problem of sketch recognition as well. 

A GRU network learns to map an input sequence $\mathbf{X} = (\mathbf{x_1,x_2}\ldots \mathbf{x_N})$ to an output sequence $\mathbf{Y} = (\mathbf{y_1,y_2}\ldots \mathbf{y_N})$. This mapping is performed by the following transformations which are applied at each time step:

\begin{eqnarray}
r_t = \sigma (W_{xr} \mathbf{x_t} + W_{hr} h_{t-1} + b_r) \\
z_t = \sigma (W_{xz} \mathbf{x_t} + W_{hz} h_{t-1} + b_z) \\
\widetilde{h_t} = tanh ( W_{xh} \mathbf{x_t} + U (r_t \odot h_{t-1}) + b_h ) \\ 
h_t = (1-z_t) \odot h_{t-1} + z_t \odot \widetilde{h_t} \\
\mathbf{y_t} = W_{hy} h_t  
\end{eqnarray}

Here, $\mathbf{x_t}$ and $\mathbf{y_t}$ represent the $t$-th input and $t$-th output respectively, $h$ represents the ``hidden'' sequence state of the GRU whose contents are regulated by parameterized gating units $r,z,\widetilde{h}$ and $\odot$ represents the elementwise dot-product. The subscripted $W$s, $b$s and $U$ represent the trainable parameters of the GRU. Please refer to Chung et al.~\cite{chung2015gated} for details.

\begin{figure}[!tbp]
    \centering    
		\includegraphics[width=0.9\linewidth]{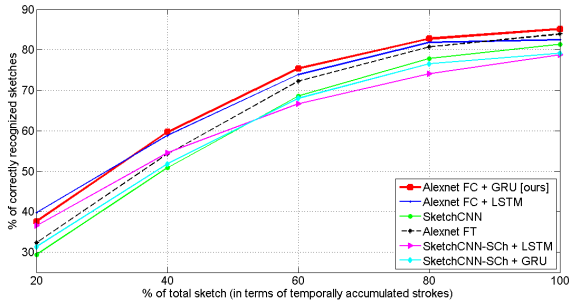}
		\caption{Comparison of online recognition performance for various classifiers. Our architecture recognizes the largest $\%$ of sketches at all levels of sketch completion. Best viewed in color.} 
		\label{fig:onlinerec}    
\end{figure}

For each sketch, information is available at temporal stroke level. We use this to construct an image sequence $\mathcal{S} = (s_1,s_2 \ldots s_N)$ of sketch strokes cumulatively accumulated over time. Thus, $s_N$ represents the full, final object sketch and $N$ represents the number of sketch strokes or equivalently, time-steps (see Figure \ref{fig:overview}). 
To represent the stroke content for each $s_i \in \mathcal{S}$, we utilize deep features obtained when $s_i$ is provided as input to Alexnet\footnote{Specifically, we remove  classification layer and use outputs from final fully-connected layer of the resulting net as features.}~\cite{krizhevsky2012imagenet}. The resulting deep feature sequence $\mathbf{X} = (\mathbf{x_1,x_2} \ldots \mathbf{,x_N})$ forms the input sequence to GRU (see Figure \ref{fig:overview}). 
The GRU unit contains $3600$ hidden units and its output is densely connected to a final softmax layer for classification. For better generalization, we include a dropout layer before the final classification layer which tends to benefit recurrent networks having a large number of hidden units. We used a dropout of $0.5$ in our experiments.

\begin{figure*}[!htbp]
    \centering    
		\includegraphics[width=0.7\textwidth,keepaspectratio]{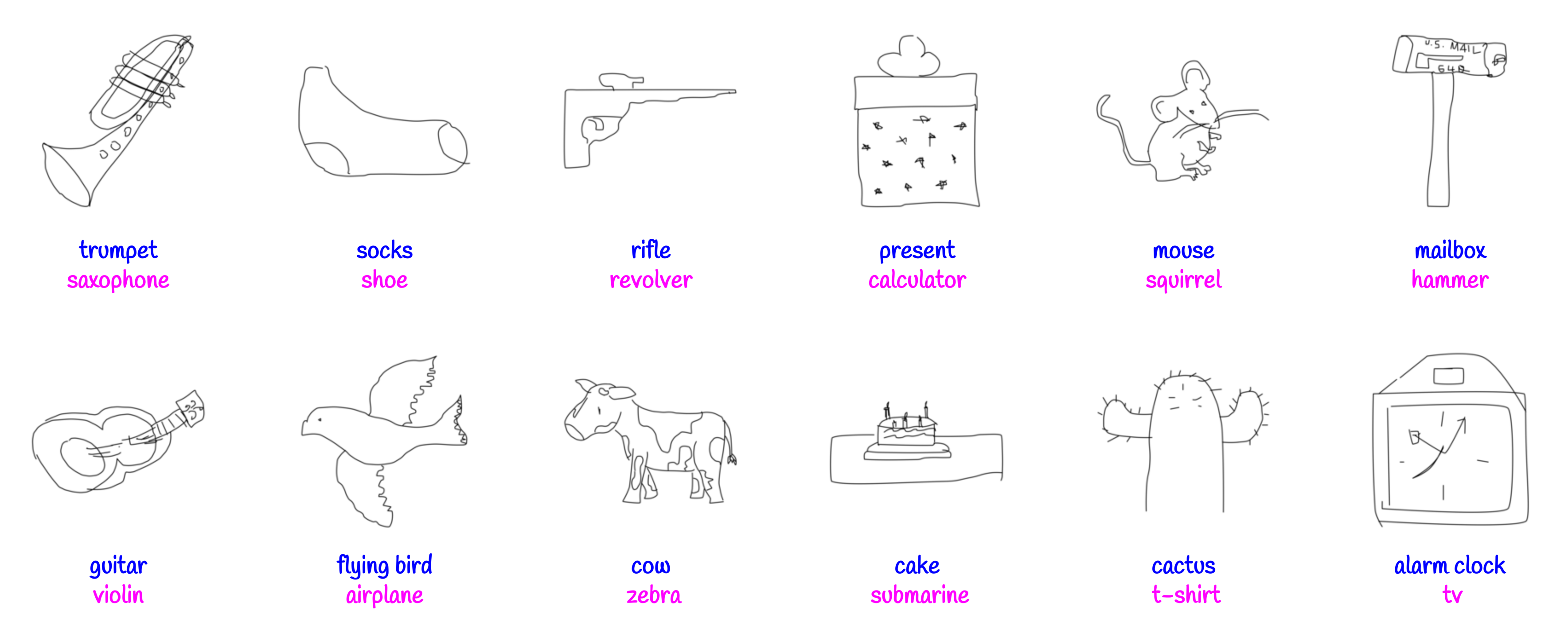}
		\caption{Examples of sketches misclassified by our framework. The ground-truth category is top label (blue) while the bottom label is our prediction (pink). Most of the misclassifications are reasonable errors.} \label{fig:semantic}    
\end{figure*}

\subsection{Training}
\label{sec:training}

Our architecture produces an output prediction $\mathbf{y_t}$ for every time-step $t, 1 \le t \le N$ in the sequence. By comparing the predictions $\mathbf{y_t}$ with the ground-truth, we can determine the corresponding loss $l_t$ for a fixed loss function (shown as a yellow box in Figure \ref{fig:overview}). This loss is weighted by a corresponding $w_{\!_{t}}$ and backpropagated) for the corresponding time step $t$. For the weighing function, we use 

\begin{equation}
w_{\!_{t}} = e^{-\alpha (1 - \frac{t}{N})} \label{eq:w}
\end{equation}

Thus, losses corresponding to final stages of sequence are weighted more to encourage correct prediction of the full sketch. Also, since $w_{\!_{t}}$ is non-zero, our design incorporates losses from all steps of the sequence. This has the net effect of encouraging correct predictions even in the early stages of the sequence. Overall, this feature enables our recognition framework to be accurate and responsive right from the beginning of the sketching process (Section \ref{sec:experiments}) in contrast with frameworks which need to wait for the sketching to finish before analysis can begin. We additionally studied variations of the weighing function given in Equation \eqref{eq:w} -- using the final sequence member loss (i.e. $w_{\!_{t}} = 0 \hspace{0.5mm} \hspace{0.5mm} \forall t \neq N$) and linearly weighted losses (i.e. $w_{\!_{t}} = \frac{t}{N}$). We found that exponentially weighted loss\footnote{We used $\alpha=10$ for our experiments.} gave superior results. 

To address the high variation of sequence length across sketches, 
%(\textit{statistics}), 
 we create batches of sketches having equal sequence length (i.e. $N$ (Sec. \ref{sec:overview})). These batches of varying size are randomly shuffled and delivered to the recurrent network during training. For each batch, categorical-cross-entropy loss is generated for each sequence by comparing the predictions with the ground-truth. The resulting losses are weighted (Equation \eqref{eq:w}) on a per-timestep basis as described previously and back-propagated through the corresponding sequence during training. We used stochastic gradient descent with a learning rate of $0.001$ for training. 

\subsection{Category prediction}
\label{sec:prediction}

Suppose for a given input sequence $\mathbf{X} = (\mathbf{x_1,x_2}$ $,\ldots\mathbf{,x_N})$, the corresponding outputs at the softmax layer are  $\mathbf{p_1,p_2,\ldots,p_N}$. Note that in our case, $\mathbf{x_t} \in \mathbb{R}^{d}$ where $d$ is the dimension of deep feature $\mathbf{x_t}$ and $\mathbf{p_t} \in \mathbb{R}^{K}$ where $K$ is the number of object categories ($160$). To determine the final category label $c_{\mathbf{X}}$, we perform a weighted sum-pooling of the softmax outputs as $c_{\mathbf{X}} = \arg\max_j \sum_{t=1}^N \mathbf{p_{t}}^{j} w_{\!_{t}}$  where $w_{\!_{t}}$ is as given in Equation \eqref{eq:w} and $ 1 \le j \le K$. 
We explored various other softmax output pooling schemes -- last sequence member-based prediction ($c_{\mathbf{X}} = \arg\max_j \mathbf{p_{N}}^{j}$), max-pooling ($c_{\mathbf{X}} = \arg\max_j \displaystyle [ \max_t \mathbf{p_{t}}^{j} \displaystyle ]$), mean-pooling ($c_{\mathbf{X}} = \arg\max_j \sum_{t=1}^{N} \mathbf{p_{t}}^{j}/N$). From our validation experiments, we found weighted sum-pooling to be the best choice overall.

\begin{table}[!tbp]
\centering
\scriptsize
\begin{tabular}{|c|c|c|c|}
\hline
\textsc{CNN} & \thead{\textsc{Recurrent} \\ \textsc{Network}} & \thead{\textsc{\#Hidden}} & \thead{\textsc{Avg.}\\ \textsc{Acc}} \\
\hline
\hline 
\textbf{Alexnet-FC} & \textbf{GRU} & \textbf{3600} & \textbf{85.1\%} \\
\hline
Alexnet-FC & LSTM & 3600 & $82.5\%$ \\
\hline
SketchCNN~\cite{yang2015deep} & - & - & $81.4\%$ \\
\hline
Alexnet-FT & - & - & $83.9\%$ \\
\hline
SketchCNN-Sch-FC & LSTM & $3600$ & $78.8\%$ \\
\hline
SketchCNN-Sch-FC & GRU & $3600$ & $79.1\%$ \\
\hline
\end{tabular}
\caption{Average recognition accuracy (rightmost column) for various architectures. \textsc{\#Hidden} refers to the number of hidden units used in recurrent network. We obtain state-of-the-art results for sketch object recognition.}
\label{tab:accuracy}
\end{table}

\vspace{-3mm}

\section{Experiments}
\label{sec:experiments}

\subsection{Data}
\label{sec:data}

In addition to obtaining the best results among approaches using handcrafted features, the work of Rosalia et al.~\cite{Schneider2014} was especially instrumental in identifying a $160$-category subset of the TU Berlin dataset which could be unambiguously recognized by humans. Consequently, our experiments are based on this curated $160$-category subset of sketches. Following Rosalia et al.~\cite{Schneider2014}, we use $56$ sketches from each of the $160$ categories. To ensure principled evaluation, we split the $56$ sketches of each category randomly into sets containing $57\%$, $18\%$ and $25\%$ of sketches to be used for training, validation and testing respectively\footnote{Thus, we have $32$, $10$ and $14$ sketches from each category in the training, validation and test sets respectively.}. Additionally, we utilized the validation set exclusively for making choices related to architecture and parameter settings and performed a one-shot comparative evaluation of ours and competing approaches on the test set.

\subsection{Other architectures}

We compared our performance with the following architectures:

\textbf{Alexnet-FT:} As a baseline experiment, we fine-tuned Alexnet using our $160$-category training data. To ensure sufficient data, we augmented the training data on the lines of Sarvadevabhatla et al.~\cite{Sarvadevabhatla2015}. We also used the final fully-connected $4096$-dimensional layer features as input to our recurrent architectures. We shall refer to such usage by \textsc{Alexnet-FC}.

\textbf{SketchCNN:} This is essentially the deep architecture of Yu et al.~\cite{yang2015deep} but retrained for the categories and splits mentioned in Section \ref{sec:data}. Since CNNs do not inherently store ``state", the authors construct six different sub-sequence stroke accumulation images which comprise the channels of the input representation to the CNNs. It comprises of five different CNNs, each trained for five different scaled versions of $256 \times 256$ sketches.  The last fully-connected layer's $512$-dimensional features from all the five CNNs are processed using a Bayesian fusion technique to obtain the final classification. 

For our experiments, we also concatenated the $512$ dimensional features from each scale of \textsc{SketchCNN} as the input feature to the recurrent neural network architectures that were evaluated. However, only the full $256 \times 256$ sketch was considered as the input to CNN (i.e. single-channel). For the rest of the paper, we refer to the resulting $2560$-dimensional feature as \textsc{SketchCNN-SCh-FC}.

\textbf{Recurrent architectures:} We experimented with the number of hidden units, the number of recurrent layers, the type of recurrent layers (i.e. LSTM or GRU), the training loss function (Section \ref{sec:training}) and various pooling methods for obtaining final prediction in terms of individual sequence member predictions (Section \ref{sec:prediction}).  

We built the software framework for our proposed architecture using Lasagne~\cite{dieleman2015lasagne} and Theano~\cite{Bastien-Theano-2012} libraries. We also used MatConvNet~\cite{vedaldi2015matconvnet} and Caffe~\cite{jia2014caffe} libraries for experiments related to other competing architectures.

\vspace{-2mm}

\subsection{Results}
\label{sec:results}

\textbf{Overall performance}: Table 1 summarizes the overall performance in terms of average recognition accuracy for various architectures. As can be seen, our GRU-based architecture (first row) outperforms SketchCNN by a significant margin even though it is trained on only $57\%$ of the total data. We believe our good performance stems from (a) being able to exploit the sequential information in a scalable and efficient manner via recurrent neural networks  (b) the superiority of the deep sketch features provided by Alexnet~\cite{Sarvadevabhatla2015} compared to the SketchCNN-FC features. The latter can be clearly seen when we compare the first two rows of Table 1 with the last two rows. In our case, the performance of GRU was better than that of LSTM when Alexnet features were used.  Overall, it is clear that the choice of (sketch) features and the recurrent network \textit{both} play a crucial role in obtaining state-of-the-art performance for the sketch recognition task. 

\textbf{On-line recognition:} We also compared the various architectures for their ability to recognize  sketches as they are being drawn (i.e. on-line recognition performance). For each classifier, we determined the fraction of test sketches $t_x$ correctly recognized when only the first $x\%$ of the temporal sketch strokes are available. We varied $x$ between $20$ to $100$ in steps of $20$ and plotted $t_x$ as a function of $x$. The results can be seen in Figure \ref{fig:onlinerec}. Intuitively, the higher a curve on the plot, the better its online recognition ability. As can be seen, our framework consistently recognizes a larger fraction of sketches at all levels of sketch completion (except for very small $x$) relative to other architectures. 

\textbf{Semantic information:} To determine the extent to which our architecture captures semantic information, we examined the performance of the classifier on misclassified sketches. As can be seen in Figure \ref{fig:semantic}, most of the misclassifications are reasonable errors (e.g. \texttt{guitar} is mistaken for \texttt{violin}) and demonstrate that our framework learns the overall semantics of the object recognition problem. 

\vspace{-4mm}

\section{Conclusion}

In this paper, we have presented our deep recurrent neural network architecture for freehand sketch recognition. Our architecture has two prominent traits. \textit{Firstly}, its design accounts for the inherently sequential and cumulative nature of human sketching process in a natural manner. \textit{Secondly}, it exploits long-term sequential and structural regularities in stroke data represented as deep features. These two traits enable our system to achieve state-of-the-art recognition results on a large database of freehand object sketches. We have also shown that our recognition framework is highly suitable for on-the-fly interpretation of sketches as they are being drawn. Our framework source-code and associated data (pre-trained models) can be accessed at  \url{https://github.com/val-iisc/sketch-obj-rec}.

%\vspace{-1mm}

\section{Acknowledgements}
We thank NVIDIA for their grant of Tesla K40 GPU.

\bibliographystyle{abbrv}

\balancecolumns

\end{document}